\pdfoutput=1

\documentclass[11pt]{article}

\usepackage[final]{acl}

\usepackage{times}
\usepackage{latexsym}
\usepackage{amsmath}

\usepackage[T1]{fontenc}

\usepackage[utf8]{inputenc}

\usepackage{microtype}

\usepackage{inconsolata}

\usepackage{graphicx}
\usepackage{makecell}

\usepackage{multirow}
\usepackage{booktabs}
\usepackage{tabularx}
\usepackage{enumitem}

\title{Think Before You Prune: Selective Self-Generated Calibration for Pruning Large Reasoning Models}

\author{
  Yang Xiang, 
  Yixin Ji, 
  Juntao Li\thanks{\; Corresponding author.}, 
  \textbf{Min Zhang} \\
  Soochow University \\
  \texttt{\{baldwin021129, jiyixin169\}@gmail.com}, \texttt{\{ljt, minzhang\}@suda.edu.cn}
}

\begin{document}
\maketitle
\begin{abstract}
Large Reasoning Models (LRMs) have demonstrated remarkable performance on complex reasoning benchmarks.
However, their long chain-of-thought reasoning processes incur significant inference overhead.
Pruning has emerged as a promising approach to reducing computational costs.
However, existing efforts have primarily focused on large language models (LLMs), while pruning LRMs remains unexplored.
In this work, we conduct the first empirical study on pruning LRMs and show that directly applying existing pruning techniques fails to yield satisfactory results.
Our findings indicate that using self-generated reasoning data for calibration can substantially improve pruning performance.
We further investigate how the difficulty and length of reasoning data affect pruning outcomes.
Our analysis reveals that challenging and moderately long self-generated reasoning data serve as ideal calibration data.
Based on these insights, we propose a Selective Self-Generated Reasoning (SSGR) data construction strategy to provide effective calibration data for pruning LRMs.
Experimental results on the DeepSeek-R1-Distill model series validate that our strategy improves the reasoning ability of pruned LRMs by 10–13\% compared to general pruning methods.
\end{abstract}

\section{Introduction}

Recently, Large Reasoning Models (LRMs) such as OpenAI's o1/o3~\cite{openai2024o1,openai2024o3} and the DeepSeek-R1~\cite{guo2025deepseek} series have achieved remarkable performance on challenging reasoning benchmarks, especially in competitive mathematics, programming, and science question answering.
Unlike traditional Large language models (LLMs), these LRMs, after undergoing large-scale reinforcement learning, have learned to perform effective deep thinking through chain-of-thought (CoT) before producing the final answer~\cite{chen2025towards}.
However, the long CoT reasoning of LRMs significantly increases inference overhead~\cite{ma2025cot}.
Striking a balance between accuracy and efficiency remains an open research question~\citep{chen2025think23overthinkingo1like,luo2025o1prunerlengthharmonizingfinetuningo1like,wang2025acceleratinglargelanguagemodel,zhang2025lightthinkerthinkingstepbystepcompression}.

\begin{figure}[t]
	\centering
    \includegraphics[width=.45\textwidth]{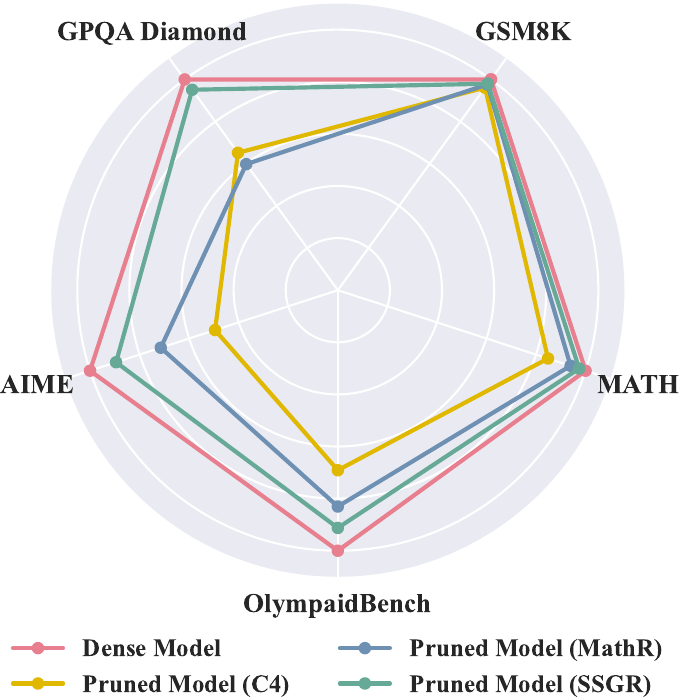}
    \caption{Performance of the pruned DeepSeek-R1-Distill-Qwen-7B model using SparseGPT with different calibration data (C4, MathR, and SSGR) across five reasoning benchmarks: GSM8K, MATH500, OlympiadBench, AIME, and GPQA Diamond.}
  \label{fig:leida}
\end{figure}

In contrast to shortening CoTs at inference time, reducing model parameters offers a more fundamental and general solution to improving efficiency.
Among various methods, pruning stands out as a promising strategy, which reduces model latency and computational cost by eliminating redundant parameters while preserving performance, and has been widely applied to LLMs~\cite{zheng2024learn,li2024adaptive,li2025t}.
Traditional magnitude-based pruning methods perform poorly on LLMs~\cite{frantar2023sparsegpt}.
Most existing methods~\cite{frantar2023sparsegpt,sun2024a,gao2024displlm} estimate parameter importance using a limited amount of data, referred to as calibration data.
The latest pruning methods~\cite{zhou2025large,li2025t} have achieved nearly lossless pruning performance.
However, these approaches are proposed for LLMs, and their applicability to LRMs remains uninvestigated.

In this work, we present the first empirical study on pruning LRMs, revealing that directly applying existing LLM pruning methods can lead to substantial degradation in LRMs' reasoning performance.
Surprisingly, we find that simply altering the calibration data can lead to dramatic improvements.
Previous studies~\cite{williams2024impact,bandari2024c4,kong2025sampleawareadaptivestructuredpruning} have shown that the choice of calibration data significantly affects performance on downstream tasks.
\citet{shin2024rethinking} observe that minimizing reconstruction error is not always optimal and can even lead to overfitting on the calibration data.
They recommend using self-generated calibration data to mitigate this issue.
Subsequently, \citet{ji2025beware} and \citet{williams2024self} propose their self-generated calibration data strategies, resulting in improved pruning performance.
However, these studies have been conducted only on LLMs, and the impact of calibration data on pruning LRMs remains unknown.

To comprehensively investigate the impact of calibration data on pruning LRMs, we design experiments to explore the impact of different types of calibration data on pruning performance.
Our empirical results indicate that choosing self-generated reasoning data as calibration data achieves the optimal pruning performance for LRMs.
Furthermore, we investigate the effects of data difficulty and reasoning length on pruning performance, finding that challenging reasoning data with moderate length serves as the ideal calibration data.
Building on this, we propose a Selective Self-Generated Reasoning (SSGR) data construction strategy that provides effective calibration data for pruning LRMs, aiming to minimize the degradation in reasoning performance after pruning.
To evaluate the effectiveness of our proposed calibration data construction method, we conduct experiments on the DeepSeek-R1-Distill model series.
As shown in Figure~\ref{fig:leida}, the results demonstrate that our method significantly outperforms baseline calibration data, offering valuable insights and guidance for pruning LRMs.

Overall, our contributions are as follows:
\begin{itemize} [leftmargin=*]
\setlength{\itemsep}{0pt}
\setlength{\parskip}{0pt}
    \item We conduct the first empirical study on pruning LRMs and find that conventional pre-training data is no longer suitable as calibration data, while self-generated reasoning data emerges as the optimal choice.

    \item Further exploration of self-generated reasoning data reveals that challenging and moderately long reasoning samples are more effective. Based on this, we propose a Selective Self-Generated Reasoning (SSGR) data construction strategy.

    \item Extensive experiments demonstrate that our SSGR data consistently outperforms baseline calibration data in pruning performance.

\end{itemize}

\section{Related work}

\paragraph{Large Reasoning Models}
OpenAI o1 series models~\cite{openai2024o1}, which are trained through large-scale reinforcement learning, perform long internal CoTs for thinking~\cite{jaech2024openai} before generating responses, achieving outstanding performance in complex reasoning tasks spanning mathematics, coding, science, and beyond.
The success of o1 marks a pivotal shift from LLMs to LRMs.
Many studies~\cite{sky_t1_2025,team2025kimi,seed2025seed,wen2025light} have made great efforts to replicate the performance of o1.
DeepSeek-R1~\cite{guo2025deepseek} achieves state-of-the-art performance by integrating cold-start data and multi-stage training.
QwQ~\cite{qwq32b} delivers comparable performance with just 32B parameters, further highlighting the effectiveness of scaling reinforcement learning in training LRMs.
Additionally, DeepSeek leverages distilled data from DeepSeek-R1 to perform supervised fine-tuning of small models (Llama and Qwen model series), also enabling test-time scaling.
Several open-source studies \cite{muennighoff2025s1,ye2025limo} have demonstrated that even a small amount of high-quality distilled data is enough to train competitive LRMs.

\paragraph{LLM Pruning}
LLM pruning can be broadly categorized into unstructured pruning~\cite{frantar2023sparsegpt,sun2024a,yin2024outlier,ilin2025thanos} and structured pruning~\cite{ma2023llm,gao2024displlm,sengupta2025you,li2025t} approaches.
We primarily focus on the former.
SparseGPT~\cite{frantar2023sparsegpt} was the first to conduct experiments on LLMs, revealing that traditional magnitude-based pruning becomes ineffective at this scale.
It introduces the use of second-order gradient information to assess parameter importance.
Wanda~\cite{sun2024a} proposes a simple and effective approach that estimates weight importance using the product of the weight magnitude and the norm of the corresponding input activations, without any retraining or weight updates.
RIA~\cite{zhang2024plugandplay} integrates weight, input, and output activations as a pruning metric.
Recent works~\cite{yin2024outlier,lu2024alphapruning,cunegatti2025zerothorder} have focused on allocating layer-wise pruning ratios instead of applying a uniform sparsity ratio across all layers.
OWL~\cite{yin2024outlier} introduces outlier-weighted layer-wise sparsity, which proportionally links each layer's sparsity to the observed outliers.
Evopress~\cite{sieberling2024evopress} leverages evolutionary algorithms to implement a dynamic compression framework for LLMs, while NEURONAL~\cite{cunegatti2025zerothorder} adaptively selects the optimal hyperparameters for block-wise and row-wise sparsity ratios.

\paragraph{Calibration data}
Calibration data is a critical component of pruning methods.
\citet{williams2024impact} find that post-training pruning is sensitive to calibration data, which significantly affects downstream performance.
\citet{shin2024rethinking} observe that the reconstruction error objective causes overfitting on calibration data and suggest using self-generated calibration data to mitigate this issue.
\citet{bandari2024c4} points out that the commonly used C4 dataset is not optimal. Arithmetic datasets are more effective as calibration data than pre-training datasets, and calibration data in the ICL format tends to have broader applicability.
\citet{ji2025beware} demonstrate that calibration data more similar to the training distribution yields better pruning performance and proposes a self-generated calibration data synthesis strategy.
Our work presents the first investigation into the impact of calibration data on pruning LRMs.
We further examine how data difficulty and reasoning length influence pruning performance and propose an effective calibration data construction strategy.


\section{The Impact of Calibration Data for Pruning LRMs.}

\begin{figure*}[t]
	\centering
    \includegraphics[width=\textwidth]{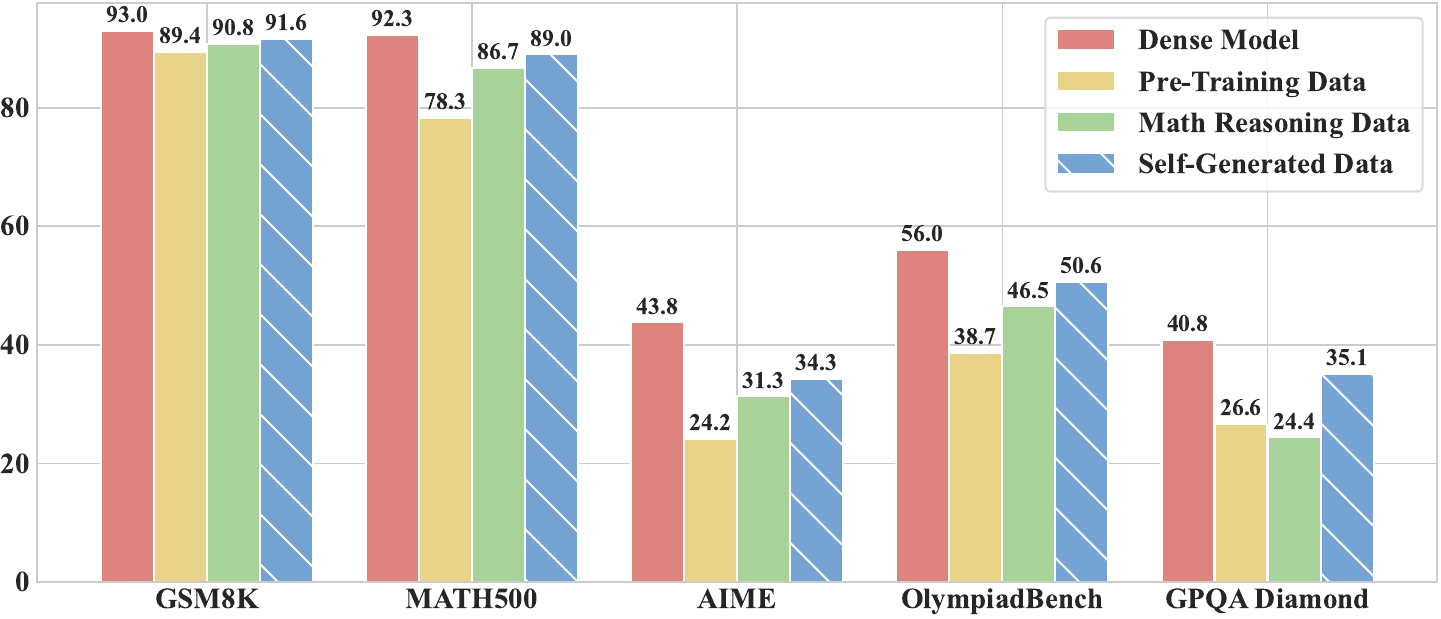}
    \caption{Comparison of pruning performance using three different types of calibration data for the dense model. For pre-training data (C4, Wikipedia, and DCLM), we report the highest accuracy among the three on each benchmark.}
  \label{fig:impact}
\end{figure*}


\subsection{Expermental Details}
\label{sec:impact}

\paragraph{Dense Model and Pruning Method}
We adopt the powerful open-source reasoning model DeepSeek-R1-Distill-Qwen-7B~\cite{guo2025deepseek} as our dense model.
For pruning, we use SparseGPT~\cite{frantar2023sparsegpt}, a widely used and effective method.
Experiments are conducted under a 50\% unstructured sparsity setting.

\paragraph{Calibration Data}
We follow prior works and randomly sample 128 sequences as calibration data.
To further highlight the impact of different calibration data on the pruned model, We extend the length of each sequence to 16,384 tokens, whereas it is typically set to 2,048 tokens.
We sample calibration data from different types of data, including:

\textbullet\ \textbf{Pre-training Data.} 
It is typically considered a primary source of calibration data.
Following \citet{ji2025beware}, We select three commonly used pre-training datasets: C4~\cite{raffel2020exploring}, Wikipedia, and DCLM~\cite{li2024datacomplm}.

\textbullet\ \textbf{Math Reasoning Data.}
The goal of pruning LRMs is to reduce the loss of reasoning capabilities in pruned model.
A straightforward idea is to use reasoning data as calibration data.
Therefore, we select GSM8K~\cite{cobbe2021training} and MATH~\cite{hendrycks2021measuring}, two widely used mathematical reasoning datasets, as the reasoning data source.
GSM8K is a high-quality elementary school math word problem dataset, while MATH is a challenging high school-level competition math problem dataset, covering seven subjects and five difficulty levels.
Each problem in both datasets includes a complete step-by-step solution. We select their training sets and combine them into the Mathematical Reasoning (MathR) Dataset.

\textbullet\ \textbf{Self-Generated Reasoning Data.}
Previous work~\cite{ji2025beware} in LLM pruning has shown that using self-generated calibration data yields better results.
We aim to explore and compare this approach in LRM pruning.
For each problem in the MathR dataset, we sample 16 responses using the dense model with a temperature of 0.6, top-p of 0.95, and a maximum generation length of 32768 tokens.
We discard responses that are overly repetitive, lack proper reasoning, or fail to follow the specified answer format.
As a result, we obtain a new dataset, referred to as the Self-Generated Reasoning (SGR) Dataset.
Apart from the experiments in Section~\ref{sec:proper}, we adopt the first correct response for each problem as the final solution.

\paragraph{Evaluation}
We evaluate the performance of the pruned models on several reasoning benchmarks, including GSM8K~\cite{cobbe2021training}, MATH500~\cite{lightman2024lets}, AIME, which contains 60 problems from the 2024 and 2025 AIME competitions, and OlympiadBench~\cite{he2024olympiadbench}.
In addition, we evaluate the pruned model’s reasoning generalization ability beyond mathematics using GPQA Diamond~\cite{rein2024gpqa}, a benchmark consisting of 198 PhD-level science questions spanning Biology, Chemistry, and Physics.
All evaluations are conducted in a zero-shot CoT setting.
To reduce output truncation and provide sufficient room for complex reasoning, we set the maximum output length to 32768 tokens.
The sampling temperature and top-p are set to 0.6 and 0.95, respectively.
We generate four responses per problem for all benchmarks and report the average pass@1 as the evaluation metric.
Our evaluation framework is built upon the open-source codebase of DeepScaleR~\cite{deepscaler2025}.

\subsection{Self-Generated Reasoning Data: A Superior Choice for Calibration}

We prune the dense model using three types of calibration data, and the performance of the pruned models is shown in Figure~\ref{fig:impact}.

\textbf{Pre-training data is not suited as calibration data for pruning LRMs.}
The pruned model using pre-training data for calibration achieves 89.4\% on GSM8K, with only a 3.3\% drop in performance.
This indicates that using pre-training data yields relatively small loss on straightforward reasoning tasks.
However, on more challenging benchmarks, the pruned model suffers a significant performance degradation, with accuracy dropping by over 10\% across the board.
Notably, the pruned model attains only 24.2\% on AIME, suggesting a substantial loss in complex reasoning capability.
As shown in Table~\ref{tab:table3}, Variation exists across different pre-training datasets.
Using C4 as calibration data achieves the best overall performance, whereas Wikipedia leads to the lowest, falling 5.6\% behind C4 on average.
DCLM surpasses C4 on AIME and matches its overall performance.


\textbf{Sampling calibration data from math reasoning data helps preserve more mathematical reasoning capability.}
Compared to pre-training data, using math reasoning data for calibration yields a 7.1\%–8.4\% improvement in accuracy on complex mathematical reasoning tasks across MATH500, AIME, and OlympiadBench.
This benefit can be attributed to the presence of explicit reasoning steps in the data, which help the model retain more mathematical reasoning capability during pruning.
However, the performance on GPQA Diamond, a scientific benchmark spanning multiple disciplines, drops significantly, falling even 2.2\% below that achieved when using pre-training data for pruning.
This suggests that while effective for mathematical tasks, math reasoning data lacks generalization and fails to transfer reasoning ability across domains.

\textbf{Self-generated reasoning data stands out as the optimal choice of calibration data for pruning LRMs.}
Using self-generated reasoning data for calibration consistently yields the best performance across all benchmarks, with the exception of AIME, where the accuracy drop remains within 6\%.
It not only performs well on mathematical tasks, but also demonstrates strong cross-domain generalization.
On GPQA Diamond, it achieves 35.1\%, outperforming Math Reasoning Data by 10.7\%.
These results clearly demonstrate that self-generated reasoning data can further bridge the gap between the pruned and original model, effectively preserving most of the model’s reasoning capability.

The key characteristic of LRMs is to produce a long thinking process before arriving at an answer.
Each response in the self-generated reasoning dataset similarly includes this thinking process.
We find that if we remove the thinking process from the response and use the modified response for calibration, the pruning performance drops significantly, showing no improvement compared to the results obtained with MathR data, as shown in Table~\ref{tab:table3}.
This suggests that the thinking process in the response is essential for use as calibration data.
The self-generated “think-then-answer” reasoning format significantly enhances pruning performance, allowing the pruned model to retain a problem-solving pattern that more closely resembles that of the original model.

\subsection{Challenging Calibration Data Enhances Pruning Performance}

During model training, challenging examples are often used to enhance reasoning capabilities~\cite{ye2025limo}. 
But does a similar principle apply to calibration data for pruning LRMs?
Therefore, we propose a reasonable hypothesis: using challenging reasoning data for calibration can lead to better performance in pruned models.

\begin{figure}[t]
	\centering
    \includegraphics[width=.45\textwidth]{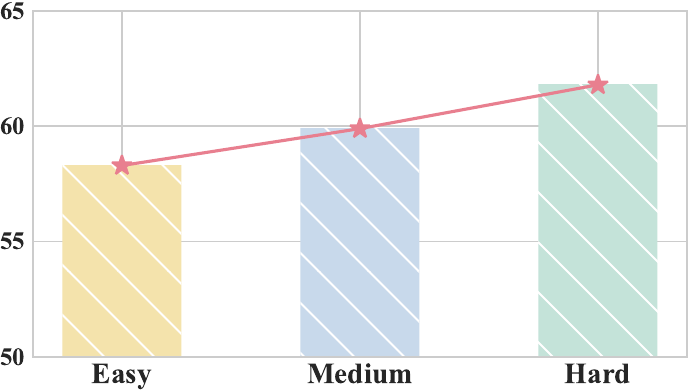}
    \caption{Comparison of Pruning Performance Using Calibration Data of Varying Difficulty Levels (\textit{Easy}, \textit{Medium}, and \textit{Hard}).}
  \label{fig:figure2}
\end{figure}

To validate this hypothesis, we first need to categorize the data based on difficulty.
The most straightforward approach is to rely on the difficulty labels provided within the datasets themselves.
For instance, GSM8K consists of elementary school math problems and is undoubtedly the simplest. 
In the MATH dataset, each problem is assigned a difficulty level, determined from a human perspective: the easiest are labeled as Level 1, while the most difficult are labeled as Level 5.
Based on these labels, we can perform a rough division of the data by difficulty.
However, such a human-defined classification is inadequate and unreliable. 
There is a significant gap between what humans and models perceive as difficult.
Moreover, different models may have entirely different notions of difficulty.
Problems that a strong model can easily solve may be highly challenging for a weaker one.
Therefore, the difficulty of the data should be determined by the model itself. 
We define a question's difficulty based on the accuracy of responses generated by the dense model.
Specifically, questions that are always answered correctly (i.e., 100\% accuracy) are labeled as \textit{Easy}; those with a majority of correct responses (i.e., accuracy > 50\%) are considered \textit{Medium}; and those with fewer correct responses (i.e., accuracy $\le$ 50\%) are classified as \textit{Hard}.

The average pruning performance using calibration data of varying difficulty levels is shown in Figure~\ref{fig:figure2}.
More details regarding the performance of different difficulty levels can be found in the Table~\ref{tab:table4}.
\textbf{As the difficulty of the calibration data increases, the performance of the pruned model improves accordingly.}
Specifically, on GSM8K and MATH500, even easy calibration data yields strong results.
However, for more challenging benchmarks like AIME and GPQA, using hard calibration data leads to significant gains, improving performance by 6.3\% and 8.6\%, respectively, compared to using easy calibration data.
This highlights the potential of letting the model identify and use challenging examples as calibration data, offering a practical and effective path toward better pruning performance.

\subsection{Proper Reasoning Length Promotes Effective Pruning}
\label{sec:proper}

\begin{table}[t]
\centering
\resizebox{0.49\textwidth}{!}{
\begin{tabular}{lcccc}
\toprule
        & \textbf{AIME} & \textbf{Olympiad} & \textbf{GPQA} & \textbf{AVG}\\
        \midrule
        \textit{Correctness} \\
        True Sample &  \textbf{36.7} & \textbf{52.1} & 35.0 &  \textbf{41.3}\\
        False Sample & 35.4 & 51.0 &  \textbf{35.4} & 40.6\\
        \midrule
        \textit{Reasoning Length} \\
        Random & 36.7 & \textbf{52.1} & 35.0 & 41.3 \\
        Short & 33.3 & 50.8 & 34.6 & 39.6 \\
        Meidum & \textbf{40.4} & 51.1 & \textbf{35.2} & \textbf{42.2} \\
        Long & 32.9 & 50.6 & 37.8 & 40.4 \\
        \bottomrule
\end{tabular}
}
\caption{Impact of Sample Correctness and Reasoning Length on Pruning Performance.}
\label{tab:table1}
\end{table}

It can be observed from Table~\ref{tab:table4} that more challenging samples generally exhibit longer reasoning lengths.
This naturally leads to the question of whether extending the reasoning length of calibration data can further enhance pruning performance.

In practice, the lengths of reasoning vary significantly among different responses to the same question.
By selecting different responses, it is possible to construct calibration datasets with varying reasoning lengths.
However, for difficult questions, the set of responses often includes incorrect answers.
Consequently, we first analyze whether incorrect answers can also serve as effective calibration data.
Following prior practice, we select the first correct response as the true sample and the first incorrect response as the false sample.
These are then used separately as calibration data for pruning.
As shown in Table~\ref{tab:table1}, pruning with incorrect samples results in lower performance compared to using correct samples.
Since they are generated by the dense model, incorrect responses still contribute positively to pruning, indicating that self-generated reasoning data, regardless of correctness, can be useful for pruning.
However, since incorrect responses fail to reach the correct final answer, they inherently contain flawed reasoning paths and introduce a certain level of noise.
As a result, the overall quality of these samples is inherently lower compared to that of correct responses.

Among the correct responses, we sort them based on their token lengths and select the shortest, median, and longest ones to construct three reasoning datasets of varying lengths, denoted as \textit{Short}, \textit{Medium}, and \textit{Long}, respectively.
Additionally, we include \textit{Random}, which randomly selects samples from the correct responses.
As shown in Table~\ref{tab:table1}, longer reasoning does not necessarily lead to better pruning performance.
The Medium sample outperforms both Short and Long by 2.6\% and 1.8\% on average, respectively.
This indicates that \textbf{moderately long reasoning tends to serve as the most effective calibration data, while overly short or excessively long responses may degrade pruning performance.}
A proper reasoning length strikes a balance: it captures sufficient intermediate thinking steps without introducing unnecessary verbosity, thus yielding better results.

\section{Method}
Based on the empirical study presented in Section 3, we propose a calibration data construction strategy for pruning LRMs.
\citet{liu2025quantization} observed that channel magnitudes and activation distributions exhibit highly similar patterns across reasoning datasets from different domains (e.g., math, code).
Therefore, we consider that using a single domain is sufficient for constructing calibration data.
Specifically, for a problem $q$ from data source $\mathcal{D}$, we perform $n$ independent sampling runs using the dense model,
resulting in a response set
$\mathcal{R} = \left \{ r_1,r_2,r_3,..,r_n \right \} $.
Let $k$ denote the number of correct responses in $\mathcal{R}$. We compute the correctness ratio as $c = \frac{k}{n}$.
If $c > \tau$, where $\tau$ is a difficulty threshold, we discard the problem $q$ from the dataset.
If $c \le  \tau $, we retain $q$ and proceed to filter $\mathcal{R}$.
We first remove incorrect responses based on exact answer matching or pattern verification.
Then, we further filter out low-quality responses, such as those that are repetitive, lack coherent reasoning, or do not conform to the expected answer format, resulting in a refined response set $\mathcal{R} ^{'} \subseteq \mathcal{R}$.
From the filtered set $\mathcal{R} ^{'}$, we select the response with the median token length as the final target response for $q$.
Formally, let $\mathcal{R'} = \left \{ r_1, r_2, ..., r_{n'} \right \} $ be sorted such that their token lengths satisfy $l_1\le l_2\le \cdot \cdot \cdot \le l_{n'}$, where $l_i$ = length($r_i$).
We define the median response as $r_{median} = r_m$, where $m = \left \lfloor \frac{n'}{2}  \right \rfloor $.
Let the total calibration data length be $l$, and the length of question $q$ be $l_q$.  
We define the available length for response as $l' = l - l_q$.
The final target response $\hat{r}$ for question $q$ is selected as:
\begin{align}
    \hat{r}=\left\{\begin{array}{ll}
    r_{median}, & \text { if } \ell_{m} \leq l', \\
    \arg \max _{\substack{r_{i} \in \mathcal{R}^{\prime} \\
    \ell_{i} \leq l}} \ell_{i}, & \text { if } \ell_{m}>l' \text { and } \exists \ell_{i} \leq l', \\
    \arg \min _{r_{i} \in \mathcal{R}^{\prime}} \ell_{i}, & otherwise .
    \end{array}\right.
\end{align}
When the median-length response is shorter than or equal to $l'$, it is chosen directly.
If the median response is longer than $l'$, we select the longest response within the length limit $l'$.
If no response is shorter than or equal to $l'$, we fall back to the shortest response available.
This strategy balances verbosity and conciseness, and avoids selecting responses that are either too short to reflect reasoning or too long to be practical.
The resulting set of problem–response pairs constitutes our selective self-generated reasoning (SSGR) calibration data.

\begin{table*}[t]
\centering
\small
\begin{tabular}{lllcccccc}
\toprule
\textbf{Model} & \textbf{Method} & \textbf{Data} & \textbf{GSM8K} & \textbf{MATH500} & \textbf{AIME} & \textbf{Olympiad} & \textbf{GPQA} & \textbf{AVG} \\
\midrule

\multirow{9}{*}{\textit{\makecell[l]{DeepSeek-\\R1-Distill-\\Qwen-7B}}}
& Dense      &        & 93.0 & 92.3 & 43.8 & 56.0 & 40.8 & 65.2 \\
\cmidrule(l){2-9}
& \multirow{4}{*}{SparseGPT}   
  & C4     & 89.4 & 78.3 & 21.7 & 38.7 & 26.6 & 50.9 \\
& & MathR & 90.8 & 86.7 & 31.3 & 46.5 & 24.4 & 55.9 \\
& & SGR   & \textbf{91.6} & 89.0 & 34.3 & 50.6 & 34.8 & 60.1 \\
& & SSGR  & 91.1 & \textbf{89.9} & \textbf{39.2} & \textbf{51.1} & \textbf{38.8} & \textbf{62.0} \\
\cmidrule(l){2-9}
& \multirow{4}{*}{Wanda}
  & C4     & 88.1 & 73.5 & 13.3 & 35.1 & 16.9 & 45.4 \\
& & MathR & 90.0 & 84.6 & 27.5 & 45.5 & 26.9 & 54.9 \\
& & SGR   & \textbf{90.5} & 86.3 & 30.0 & 46.3 & 30.1 & 56.6 \\
& & SSGR  & 90.4 & \textbf{87.3} & \textbf{32.5} & \textbf{48.3} & \textbf{31.7} & \textbf{58.0} \\
\midrule

\multirow{9}{*}{\textit{\makecell[l]{DeepSeek-\\R1-Distill-\\Llama-8B}}}
& Dense   & & 91.6 & 90.3 & 45.4 & 54.4 & 42.4 & 64.8 \\
\cmidrule(l){2-9}
& \multirow{4}{*}{SparseGPT}   
  & C4     & 76.4 & 56.0 & 5.8 & 27.7 & 14.1 & 36.0 \\
& & MathR & 80.4 & 70.6 & 19.6 & 38.8 & 9.5 & 43.8     \\
& & SGR   & 84.2 & 76.0 & 22.5 & 41.2 & 17.3 & 48.2   \\
& & SSGR  & \textbf{84.8} & \textbf{77.1} & \textbf{26.3} & \textbf{41.5} & \textbf{18.2} & \textbf{49.6}  \\
\cmidrule(l){2-9}
& \multirow{4}{*}{Wanda}
  & C4     & 69.0 & 44.8 & 5.0 & 21.6 & 8.6 & 29.8    \\
& & MathR &  74.8 & 58.8 & 8.3 & 27.7 & 9.6 & 35.8    \\
& & SGR   &  76.7 & 61.8 & \textbf{10.0} & 28.6 & \textbf{11.4} & 37.7    \\
& & SSGR  &  \textbf{76.8} & \textbf{63.4} & \textbf{10.0} & \textbf{29.9} & 10.9 & \textbf{38.2} \\
           
\bottomrule
\end{tabular}
\caption{Pruning performance under a 50\% sparsity ratio with different calibration data on DeepSeek-R1-Distill-Qwen-7B and DeepSeek-R1-Distill-Llama-8B. MathR, SGR, and SSGR refer to Math Reasoning, Self-Generated Reasoning, and our Selective Self-Generated Reasoning Data, respectively. Olympiad and GPQA correspond to OlympiadBench and GPQA Diamond. The best results are shown in bold.
}
\label{tab:table2}
\end{table*}

\section{Experiments}

\subsection{Experimental Details}

To evaluate the effectiveness of our proposed calibration data construction method, we apply it to several LRMs, including DeepSeek-R1-Distill-Qwen-7B and DeepSeek-R1-Distill-Llama-8B.
As described in Section~\ref{sec:impact}, we use pre-training data, math reasoning data, and self-generated reasoning data as baseline sources for calibration data, with C4 selected to represent the pre-training data.
We adopt two widely used pruning methods, SparseGPT and Wanda, to prune the dense models.
In our main experiments, we report model performance under a 50\% sparsity setting.
The pruned models are evaluated on five reasoning benchmarks: GSM8K, MATH500, OlympiadBench, AIME, and GPQA.
During the construction of the selective self-generated reasoning data, MathR is chosen as the data source, and 16 responses are sampled for each problem.
We set the difficulty threshold to 0.75 to retain a broader pool of candidate samples.

\subsection{Overall Performance}
We report the main results in Table~\ref{tab:table2}.
Overall, our calibration data construction method outperforms other baseline calibration data on complex reasoning benchmarks, and is compatible with different pruning methods.
On DeepSeek-R1-Distill-Qwen-7B, the Selective Self-Generated Reasoning data (SSGR) achieves an average performance improvement of 11.9\% to 13.6\% over pre-training data (C4) and 3.1\% to 6.1\% over math reasoning data.
This clearly demonstrates the effectiveness of using self-generated reasoning data for calibration.
Moreover, SSGR brings an additional 1.4\% to 1.9\% average gain over the unfiltered Self-Generated Reasoning (SGR) data, highlighting the necessity of our reasoning data selection strategy.
On DeepSeek-R1-Distill-Llama-8B, using SSGR as the calibration data also outperforms all other baseline data.
SSGR improves performance by 8.4\% to 13.6\% over C4 and by 2.4\% to 5.8\% over MathR, and also yields a 0.5\% to 1.4\% gain compared to SGR. 
However, it is evident that DeepSeek-R1-Distill-Llama-8B experiences a more significant performance drop.When using SparseGPT with SSGR as the calibration data, it suffers a 15.2\% performance loss, whereas DeepSeek-R1-Distill-Qwen-7B only shows a 3.2\% loss.
Using Wanda results in even greater performance degradation.
This suggests that DeepSeek-R1-Distill-Llama-8B is more difficult to compress, and improvements from calibration data alone are insufficient; more effective pruning techniques are necessary to achieve better results.

The pruning performance gains from self-generated reasoning data can be attributed to the fact that the generated responses guide the model to retain similar reasoning patterns during pruning, thereby reducing the discrepancy between the pruned and dense models.
Moreover, by selecting responses with higher difficulty and appropriate length, we ensure the data better serves as effective calibration data for pruning LRMs.
In addition, although the selected calibration data is derived entirely from mathematical problems, the pruned models still exhibit strong performance on out-of-domain tasks such as GPQA, indicating the generalizability of our data construction method.

\section{Discussion}
\subsection{Is the Calibration Data Construction Method Applicable to Other Pruning Settings?}

\begin{figure}[t]
	\centering
    \includegraphics[width=.45\textwidth]{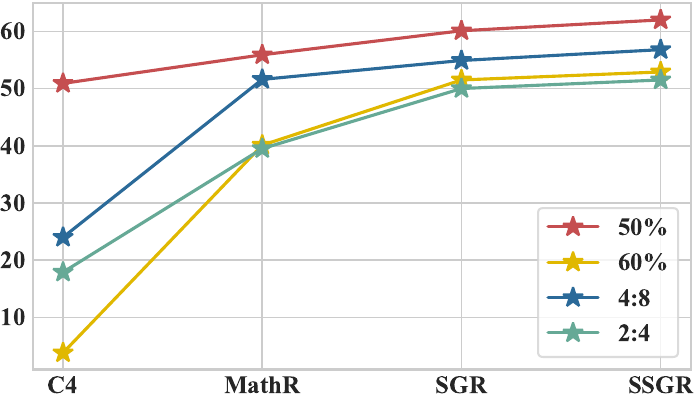}
    \caption{Pruning performance under various pruning settings with different calibration data.}
  \label{fig:more}
\end{figure}

We further validate the effectiveness of the selective self-generated reasoning (SSGR) data under a broader range of pruning settings.
As shown in Figure~\ref{fig:more}, when applying SparseGPT with unstructured 60\% pruning and semi-structured 4:8 and 2:4 sparsity settings, SSGR consistently outperforms all baseline calibration data.
Notably, at higher sparsity levels and under semi-structured pruning, using pre-training data for calibration leads to a sharp drop in pruning performance. This further demonstrates that pre-training data is ill-suited for calibrating pruned LLMs. In contrast, calibration with SGR data significantly mitigates this issue, yielding a 30.9\%–47.7\% improvement over C4.
Moreover, the SSGR subset derived from SGR offers an additional performance gain of 1.1\%–1.9\% over SGR.
This further demonstrates the strong generalizability of SSGR across various settings.

\subsection{How Does Pruning Affect the Generation Length of LRMs?}

\begin{figure}[t]
	\centering
    \includegraphics[width=.45\textwidth]{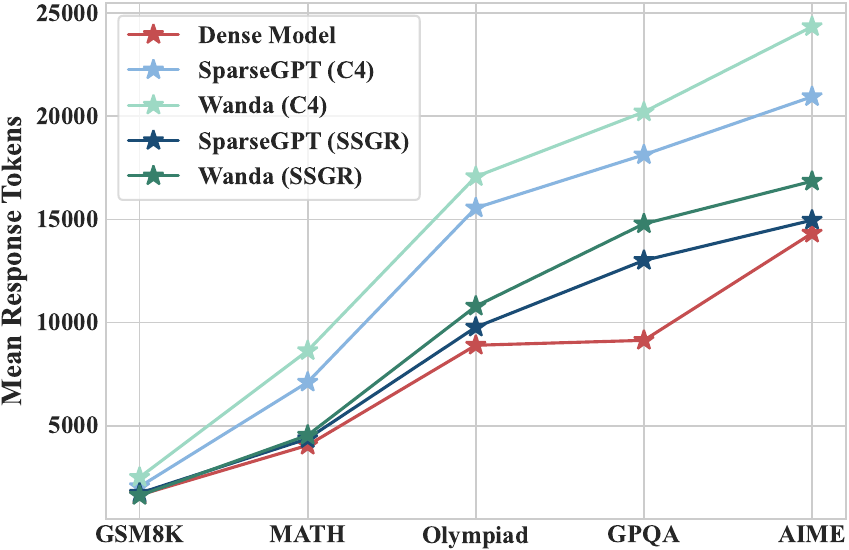}
    \caption{Average response length of pruned LRMs across different benchmarks.}
  \label{fig:Generation_Length}
\end{figure}

Reasoning length is a key factor affecting the efficiency of inference in large reasoning models.
While pruning reduces the model size by decreasing the number of parameters, its actual impact on inference speed remains uncertain.
In Figure~\ref{fig:Generation_Length}, we report average response lengths of pruned LRMs.
On simple reasoning tasks, the pruned models produce outputs nearly identical in length to those of the dense model.
However, for complex tasks, pruned models tend to generate longer responses.
Pruned LRMs calibrated with SSGR data produce significantly shorter outputs compared to those calibrated with C4.
Pruned LRMs with less performance degradation tend to generate outputs closer in length to the dense model.
Therefore, in practical applications, it is crucial to minimize the performance degradation of pruned LRMs, not only to ensure better pruning effectiveness but also to avoid generating excessively long responses that can lead to inference delays.
Recent studies~\cite{hou2025thinkprune,yi2025shorterbetter} have also explored methods to compress reasoning length.
It is worth noting that model compression and chain-of-thought (CoT) compression are orthogonal and can be applied together.
This opens up the possibility of building reasoning models that are both parameter-efficient and capable of fast inference.

\section{Conclusion}

In this work, we conduct the first empirical investigation into pruning large reasoning models.
Using pre-training data directly as calibration input for pruning results in a substantial degradation in the reasoning performance of LRMs.
In contrast, leveraging self-generated reasoning data for calibration notably enhances pruning performance.
We further observe that self-generated reasoning data with higher difficulty and moderate length serves as ideal calibration data.
Building on these insights, we propose a selective self-generated reasoning (SSGR) data construction strategy to provide effective calibration data for pruning LRMs.
Experiments on the DeepSeek-R1-Distill-Qwen and DeepSeek-R1-Distill-Llama model series show that our SSGR strategy significantly outperforms baseline calibration data, providing practical insights and guidance for pruning large reasoning models.

\section*{Limitations}
Although our proposed Selective Self-Generated Reasoning (SSGR) data construction strategy has made notable progress in pruning LRMs, there are still some limitations:
\begin{itemize}[leftmargin=*]
    \setlength{\itemsep}{0pt}
    \setlength{\parskip}{0pt}
    \item Due to computational resource constraints, we conduct experiments only on LLMs up to 14B.
    \item In our data construction process, we sample 16 responses for each question, which is time-consuming. In the future, We plan to explore methods that enable the model to generate an optimal response in a single pass, significantly improving construction efficiency.
\end{itemize}

\bibliography{custom.bib}

@misc{openai2024o1,
  author = {OpenAI},
  title = {Introducing OpenAI O1 Preview},
  year = {2024},
  url = {https://openai.com/index/introducing-openai-o1-preview}
}

@misc{openai2024o3,
  author = {OpenAI},
  title = {Openai o3-mini system card},
  year = {2025},
  url = {https://openai.com/index/o3-mini-system-card}
}

@article{ma2025cot,
  title={CoT-Valve: Length-Compressible Chain-of-Thought Tuning},
  author={Ma, Xinyin and Wan, Guangnian and Yu, Runpeng and Fang, Gongfan and Wang, Xinchao},
  journal={arXiv preprint arXiv:2502.09601},
  year={2025}
}

@article{ma2023llm,
  title={Llm-pruner: On the structural pruning of large language models},
  author={Ma, Xinyin and Fang, Gongfan and Wang, Xinchao},
  journal={Advances in neural information processing systems},
  volume={36},
  pages={21702--21720},
  year={2023}
}

@inproceedings{
gao2024displlm,
title={{DISP}-{LLM}: Dimension-Independent Structural Pruning for Large Language Models},
author={Shangqian Gao and Chi-Heng Lin and Ting Hua and Zheng Tang and Yilin Shen and Hongxia Jin and Yen-Chang Hsu},
booktitle={The Thirty-eighth Annual Conference on Neural Information Processing Systems},
year={2024},
url={https://openreview.net/forum?id=YxaY6tHgg0}
}

@inproceedings{
sengupta2025you,
title={You Only Prune Once: Designing Calibration-Free Model Compression With Policy Learning},
author={Ayan Sengupta and Siddhant Chaudhary and Tanmoy Chakraborty},
booktitle={The Thirteenth International Conference on Learning Representations},
year={2025},
url={https://openreview.net/forum?id=5RZoYIT3u6}
}

@article{zhou2025large,
  title={Large Language Model Compression with Global Rank and Sparsity Optimization},
  author={Zhou, Changhai and Qiao, Qian and Zhang, Weizhong and Jin, Cheng},
  journal={arXiv preprint arXiv:2505.03801},
  year={2025}
}

@inproceedings{
zheng2024learn,
title={Learn To be Efficient: Build Structured Sparsity in Large Language Models},
author={Haizhong Zheng and Xiaoyan Bai and Xueshen Liu and Zhuoqing Mao and Beidi Chen and Fan Lai and Atul Prakash},
booktitle={The Thirty-eighth Annual Conference on Neural Information Processing Systems},
year={2024},
url={https://openreview.net/forum?id=iSfCWhvEGA}
}

@inproceedings{
li2024adaptive,
title={Adaptive Layer Sparsity for Large Language Models via Activation Correlation Assessment},
author={Wei Li and Lujun Li and Mark G. Lee and Shengjie Sun},
booktitle={The Thirty-eighth Annual Conference on Neural Information Processing Systems},
year={2024},
url={https://openreview.net/forum?id=Jup0qZxH7U}
}

@article{li2025t,
  title={T$\backslash$'yr-the-Pruner: Unlocking Accurate 50\% Structural Pruning for LLMs via Global Sparsity Distribution Optimization},
  author={Li, Guanchen and Xu, Yixing and Li, Zeping and Liu, Ji and Yin, Xuanwu and Li, Dong and Barsoum, Emad},
  journal={arXiv preprint arXiv:2503.09657},
  year={2025}
}

@article{chen2025towards,
  title={Towards reasoning era: A survey of long chain-of-thought for reasoning large language models},
  author={Chen, Qiguang and Qin, Libo and Liu, Jinhao and Peng, Dengyun and Guan, Jiannan and Wang, Peng and Hu, Mengkang and Zhou, Yuhang and Gao, Te and Che, Wanxiang},
  journal={arXiv preprint arXiv:2503.09567},
  year={2025}
}

@article{team2025kimi,
  title={Kimi k1. 5: Scaling reinforcement learning with llms},
  author={Team, Kimi and Du, Angang and Gao, Bofei and Xing, Bowei and Jiang, Changjiu and Chen, Cheng and Li, Cheng and Xiao, Chenjun and Du, Chenzhuang and Liao, Chonghua and others},
  journal={arXiv preprint arXiv:2501.12599},
  year={2025}
}

@inproceedings{yin2024outlier,
  title={Outlier Weighed Layerwise Sparsity (OWL): A Missing Secret Sauce for Pruning LLMs to High Sparsity},
  author={Yin, Lu and Wu, You and Zhang, Zhenyu and Hsieh, Cheng-Yu and Wang, Yaqing and Jia, Yiling and Li, Gen and Jaiswal, Ajay and Pechenizkiy, Mykola and Liang, Yi and others},
  booktitle={The Forty-First International Conference on Machine Learning},
  year={2024}
}

@article{lu2024alphapruning,
  title={Alphapruning: Using heavy-tailed self regularization theory for improved layer-wise pruning of large language models},
  author={Lu, Haiquan and Zhou, Yefan and Liu, Shiwei and Wang, Zhangyang and Mahoney, Michael W and Yang, Yaoqing},
  journal={Advances in Neural Information Processing Systems},
  volume={37},
  pages={9117--9152},
  year={2024}
}

@article{ilin2025thanos,
  title={Thanos: A Block-wise Pruning Algorithm for Efficient Large Language Model Compression},
  author={Ilin, Ivan and Richtarik, Peter},
  journal={arXiv preprint arXiv:2504.05346},
  year={2025}
}

@article{sieberling2024evopress,
  title={Evopress: Towards optimal dynamic model compression via evolutionary search},
  author={Sieberling, Oliver and Kuznedelev, Denis and Kurtic, Eldar and Alistarh, Dan},
  journal={arXiv preprint arXiv:2410.14649},
  year={2024}
}

@inproceedings{
cunegatti2025zerothorder,
title={Zeroth-Order Adaptive Neuron Alignment Based Pruning without Re-Training},
author={Elia Cunegatti and Leonardo Lucio Custode and Giovanni Iacca},
booktitle={Sparsity in LLMs (SLLM): Deep Dive into Mixture of Experts, Quantization, Hardware, and Inference},
year={2025},
url={https://openreview.net/forum?id=peyLy5ek4w}
}

@article{seed2025seed,
  title={Seed-thinking-v1. 5: Advancing superb reasoning models with reinforcement learning},
  author={Seed, ByteDance and Yuan, Yufeng and Yue, Yu and Wang, Mingxuan and Zuo, Xiaochen and Chen, Jiaze and Yan, Lin and Xu, Wenyuan and Zhang, Chi and Liu, Xin and others},
  journal={arXiv preprint arXiv:2504.13914},
  year={2025}
}

@article{williams2024self,
  title={Self-calibration for Language Model Quantization and Pruning},
  author={Williams, Miles and Chrysostomou, George and Aletras, Nikolaos},
  journal={arXiv preprint arXiv:2410.17170},
  year={2024}
}

@article{guo2025deepseek,
  title={Deepseek-r1: Incentivizing reasoning capability in llms via reinforcement learning},
  author={Guo, Daya and Yang, Dejian and Zhang, Haowei and Song, Junxiao and Zhang, Ruoyu and Xu, Runxin and Zhu, Qihao and Ma, Shirong and Wang, Peiyi and Bi, Xiao and others},
  journal={arXiv preprint arXiv:2501.12948},
  year={2025}
}

@misc{qwq32b,
    title = {QwQ-32B: Embracing the Power of Reinforcement Learning},
    url = {https://qwenlm.github.io/blog/qwq-32b/},
    author = {Qwen Team},
    month = {March},
    year = {2025}
}

@article{muennighoff2025s1,
  title={s1: Simple test-time scaling},
  author={Muennighoff, Niklas and Yang, Zitong and Shi, Weijia and Li, Xiang Lisa and Fei-Fei, Li and Hajishirzi, Hannaneh and Zettlemoyer, Luke and Liang, Percy and Cand{\`e}s, Emmanuel and Hashimoto, Tatsunori},
  journal={arXiv preprint arXiv:2501.19393},
  year={2025}
}

@article{ye2025limo,
  title={LIMO: Less is More for Reasoning},
  author={Ye, Yixin and Huang, Zhen and Xiao, Yang and Chern, Ethan and Xia, Shijie and Liu, Pengfei},
  journal={arXiv preprint arXiv:2502.03387},
  year={2025}
}

@misc{deepscaler2025,
  title={DeepScaleR: Surpassing O1-Preview with a 1.5B Model by Scaling RL},
  author={Michael Luo and Sijun Tan and Justin Wong and Xiaoxiang Shi and William Y. Tang and Manan Roongta and Colin Cai and Jeffrey Luo and Tianjun Zhang and Li Erran Li and Raluca Ada Popa and Ion Stoica},
  url = {https://pretty-radio-b75.notion.site/DeepScaleR-Surpassing-O1-Preview-with-a-1-5B-Model-by-Scaling-RL-19681902c1468005bed8ca303013a4e2},
  note={Notion Blog},
  year={2025}
}

@inproceedings{
hendrycks2021measuring,
title={Measuring Mathematical Problem Solving With the {MATH} Dataset},
author={Dan Hendrycks and Collin Burns and Saurav Kadavath and Akul Arora and Steven Basart and Eric Tang and Dawn Song and Jacob Steinhardt},
booktitle={Thirty-fifth Conference on Neural Information Processing Systems Datasets and Benchmarks Track (Round 2)},
year={2021},
url={https://openreview.net/forum?id=7Bywt2mQsCe}
}

@inproceedings{he2024olympiadbench,
  title={OlympiadBench: A Challenging Benchmark for Promoting AGI with Olympiad-Level Bilingual Multimodal Scientific Problems},
  author={He, Chaoqun and Luo, Renjie and Bai, Yuzhuo and Hu, Shengding and Thai, Zhen and Shen, Junhao and Hu, Jinyi and Han, Xu and Huang, Yujie and Zhang, Yuxiang and others},
  booktitle={Proceedings of the 62nd Annual Meeting of the Association for Computational Linguistics (Volume 1: Long Papers)},
  pages={3828--3850},
  year={2024}
}

@inproceedings{
rein2024gpqa,
title={{GPQA}: A Graduate-Level Google-Proof Q\&A Benchmark},
author={David Rein and Betty Li Hou and Asa Cooper Stickland and Jackson Petty and Richard Yuanzhe Pang and Julien Dirani and Julian Michael and Samuel R. Bowman},
booktitle={First Conference on Language Modeling},
year={2024},
url={https://openreview.net/forum?id=Ti67584b98}
}

@inproceedings{frantar2023sparsegpt,
  title={Sparsegpt: Massive language models can be accurately pruned in one-shot},
  author={Frantar, Elias and Alistarh, Dan},
  booktitle={International Conference on Machine Learning},
  pages={10323--10337},
  year={2023},
  organization={PMLR}
}

@article{raffel2020exploring,
  title={Exploring the limits of transfer learning with a unified text-to-text transformer},
  author={Raffel, Colin and Shazeer, Noam and Roberts, Adam and Lee, Katherine and Narang, Sharan and Matena, Michael and Zhou, Yanqi and Li, Wei and Liu, Peter J},
  journal={Journal of machine learning research},
  volume={21},
  number={140},
  pages={1--67},
  year={2020}
}

@article{jaech2024openai,
  title={Openai o1 system card},
  author={Jaech, Aaron and Kalai, Adam and Lerer, Adam and Richardson, Adam and El-Kishky, Ahmed and Low, Aiden and Helyar, Alec and Madry, Aleksander and Beutel, Alex and Carney, Alex and others},
  journal={arXiv preprint arXiv:2412.16720},
  year={2024}
}

@misc{sky_t1_2025,
  author       = {NovaSky Team},
  title        = {Sky-T1: Train your own O1 preview model within \$450},
  howpublished = {https://novasky-ai.github.io/posts/sky-t1},
  note         = {Accessed: 2025-01-09},
  year         = {2025}
}

@article{hou2025thinkprune,
  title={Thinkprune: Pruning long chain-of-thought of llms via reinforcement learning},
  author={Hou, Bairu and Zhang, Yang and Ji, Jiabao and Liu, Yujian and Qian, Kaizhi and Andreas, Jacob and Chang, Shiyu},
  journal={arXiv preprint arXiv:2504.01296},
  year={2025}
}

@article{yi2025shorterbetter,
  title={ShorterBetter: Guiding Reasoning Models to Find Optimal Inference Length for Efficient Reasoning},
  author={Yi, Jingyang and Wang, Jiazheng},
  journal={arXiv preprint arXiv:2504.21370},
  year={2025}
}

@article{wen2025light,
  title={Light-r1: Curriculum sft, dpo and rl for long cot from scratch and beyond},
  author={Wen, Liang and Cai, Yunke and Xiao, Fenrui and He, Xin and An, Qi and Duan, Zhenyu and Du, Yimin and Liu, Junchen and Tang, Lifu and Lv, Xiaowei and others},
  journal={arXiv preprint arXiv:2503.10460},
  year={2025}
}

@inproceedings{
ji2025beware,
title={Beware of Calibration Data for Pruning Large Language Models},
author={Yixin Ji and Yang Xiang and Juntao Li and Qingrong Xia and Ping Li and Xinyu Duan and Zhefeng Wang and Min Zhang},
booktitle={The Thirteenth International Conference on Learning Representations},
year={2025},
url={https://openreview.net/forum?id=x83w6yGIWb}
}

@inproceedings{williams2024impact,
  title={On the Impact of Calibration Data in Post-training Quantization and Pruning},
  author={Williams, Miles and Aletras, Nikolaos},
  booktitle={Proceedings of the 62nd Annual Meeting of the Association for Computational Linguistics (Volume 1: Long Papers)},
  pages={10100--10118},
  year={2024}
}

@article{shin2024rethinking,
  title={Rethinking Pruning Large Language Models: Benefits and Pitfalls of Reconstruction Error Minimization},
  author={Shin, Sungbin and Park, Wonpyo and Lee, Jaeho and Lee, Namhoon},
  journal={arXiv preprint arXiv:2406.15524},
  year={2024}
}

@inproceedings{
sun2024a,
title={A Simple and Effective Pruning Approach for Large Language Models},
author={Mingjie Sun and Zhuang Liu and Anna Bair and J Zico Kolter},
booktitle={The Twelfth International Conference on Learning Representations},
year={2024},
url={https://openreview.net/forum?id=PxoFut3dWW}
}

@inproceedings{
zhang2024plugandplay,
title={Plug-and-Play: An Efficient Post-training Pruning Method for Large Language Models},
author={Yingtao Zhang and Haoli Bai and Haokun Lin and Jialin Zhao and Lu Hou and Carlo Vittorio Cannistraci},
booktitle={The Twelfth International Conference on Learning Representations},
year={2024},
url={https://openreview.net/forum?id=Tr0lPx9woF}
}

@inproceedings{bandari2024c4,
  title={Is C4 Dataset Optimal for Pruning? An Investigation of Calibration Data for LLM Pruning},
  author={Bandari, Abhinav and Yin, Lu and Hsieh, Cheng-Yu and Jaiswal, Ajay and Chen, Tianlong and Shen, Li and Krishna, Ranjay and Liu, Shiwei},
  booktitle={Proceedings of the 2024 Conference on Empirical Methods in Natural Language Processing},
  pages={18089--18099},
  year={2024}
}

@inproceedings{
li2024datacomplm,
title={DataComp-{LM}: In search of the next generation of training sets for language models},
author={Jeffrey Li and Alex Fang and Georgios Smyrnis and Maor Ivgi and Matt Jordan and Samir Yitzhak Gadre and Hritik Bansal and Etash Kumar Guha and Sedrick Keh and Kushal Arora and Saurabh Garg and Rui Xin and Niklas Muennighoff and Reinhard Heckel and Jean Mercat and Mayee F Chen and Suchin Gururangan and Mitchell Wortsman and Alon Albalak and Yonatan Bitton and Marianna Nezhurina and Amro Kamal Mohamed Abbas and Cheng-Yu Hsieh and Dhruba Ghosh and Joshua P Gardner and Maciej Kilian and Hanlin Zhang and Rulin Shao and Sarah M Pratt and Sunny Sanyal and Gabriel Ilharco and Giannis Daras and Kalyani Marathe and Aaron Gokaslan and Jieyu Zhang and Khyathi Chandu and Thao Nguyen and Igor Vasiljevic and Sham M. Kakade and Shuran Song and Sujay Sanghavi and Fartash Faghri and Sewoong Oh and Luke Zettlemoyer and Kyle Lo and Alaaeldin El-Nouby and Hadi Pouransari and Alexander T Toshev and Stephanie Wang and Dirk Groeneveld and Luca Soldaini and Pang Wei Koh and Jenia Jitsev and Thomas Kollar and Alex Dimakis and Yair Carmon and Achal Dave and Ludwig Schmidt and Vaishaal Shankar},
booktitle={The Thirty-eight Conference on Neural Information Processing Systems Datasets and Benchmarks Track},
year={2024},
url={https://openreview.net/forum?id=CNWdWn47IE}
}

@article{cobbe2021training,
  title={Training verifiers to solve math word problems},
  author={Cobbe, Karl and Kosaraju, Vineet and Bavarian, Mohammad and Chen, Mark and Jun, Heewoo and Kaiser, Lukasz and Plappert, Matthias and Tworek, Jerry and Hilton, Jacob and Nakano, Reiichiro and others},
  journal={arXiv preprint arXiv:2110.14168},
  year={2021}
}

@inproceedings{
lightman2024lets,
title={Let's Verify Step by Step},
author={Hunter Lightman and Vineet Kosaraju and Yuri Burda and Harrison Edwards and Bowen Baker and Teddy Lee and Jan Leike and John Schulman and Ilya Sutskever and Karl Cobbe},
booktitle={The Twelfth International Conference on Learning Representations},
year={2024},
url={https://openreview.net/forum?id=v8L0pN6EOi}
}

@article{liu2025quantization,
  title={Quantization hurts reasoning? an empirical study on quantized reasoning models},
  author={Liu, Ruikang and Sun, Yuxuan and Zhang, Manyi and Bai, Haoli and Yu, Xianzhi and Yu, Tiezheng and Yuan, Chun and Hou, Lu},
  journal={arXiv preprint arXiv:2504.04823},
  year={2025}
}

@misc{kong2025sampleawareadaptivestructuredpruning,
      title={Sample-aware Adaptive Structured Pruning for Large Language Models}, 
      author={Jun Kong and Xinge Ma and Jin Wang and Xuejie Zhang},
      year={2025},
      eprint={2503.06184},
      archivePrefix={arXiv},
      primaryClass={cs.CL},
      url={https://arxiv.org/abs/2503.06184}, 
}

@misc{chen2025think23overthinkingo1like,
      title={Do NOT Think That Much for 2+3=? On the Overthinking of o1-Like LLMs}, 
      author={Xingyu Chen and Jiahao Xu and Tian Liang and Zhiwei He and Jianhui Pang and Dian Yu and Linfeng Song and Qiuzhi Liu and Mengfei Zhou and Zhuosheng Zhang and Rui Wang and Zhaopeng Tu and Haitao Mi and Dong Yu},
      year={2025},
      eprint={2412.21187},
      archivePrefix={arXiv},
      primaryClass={cs.CL},
      url={https://arxiv.org/abs/2412.21187}, 
}

@misc{luo2025o1prunerlengthharmonizingfinetuningo1like,
      title={O1-Pruner: Length-Harmonizing Fine-Tuning for O1-Like Reasoning Pruning}, 
      author={Haotian Luo and Li Shen and Haiying He and Yibo Wang and Shiwei Liu and Wei Li and Naiqiang Tan and Xiaochun Cao and Dacheng Tao},
      year={2025},
      eprint={2501.12570},
      archivePrefix={arXiv},
      primaryClass={cs.CL},
      url={https://arxiv.org/abs/2501.12570}, 
}

@misc{wang2025acceleratinglargelanguagemodel,
      title={Accelerating Large Language Model Reasoning via Speculative Search}, 
      author={Zhihai Wang and Jie Wang and Jilai Pan and Xilin Xia and Huiling Zhen and Mingxuan Yuan and Jianye Hao and Feng Wu},
      year={2025},
      eprint={2505.02865},
      archivePrefix={arXiv},
      primaryClass={cs.CL},
      url={https://arxiv.org/abs/2505.02865}, 
}

@misc{zhang2025lightthinkerthinkingstepbystepcompression,
      title={LightThinker: Thinking Step-by-Step Compression}, 
      author={Jintian Zhang and Yuqi Zhu and Mengshu Sun and Yujie Luo and Shuofei Qiao and Lun Du and Da Zheng and Huajun Chen and Ningyu Zhang},
      year={2025},
      eprint={2502.15589},
      archivePrefix={arXiv},
      primaryClass={cs.CL},
      url={https://arxiv.org/abs/2502.15589}, 
}

\appendix

\begin{table*}[t]
\centering
\small
\begin{tabular}{lllcccccc}
\toprule
        \textbf{Data} & \textbf{GSM8K} & \textbf{MATH500} & \textbf{AIME} & \textbf{Olympiadbench} & \textbf{GPQA} & \textbf{AVG} \\
        \midrule
        C4 & 89.4 & 78.3 & 21.7 & 38.7 & 26.6 & 50.9 \\
        Wikipedia & 87.1 & 72.6 & 17.5 & 33.7 & 15.5 & 45.3 \\
        DCLM & 89.1 & 77.7 & 24.2 & 38.2 & 24.6 & 50.8 \\
        MathR & 90.8 & 86.7 & 31.3 & 46.5 & 24.4 & 55.9 \\
        SGR & 91.6 & 89.0 & 34.3 & 50.6 & 34.8 & 60.1 \\
        SGR \textit{w/o thinking} & 90.4 & 84.9 & 31.3 & 46.4 & 25.6 & 55.7 \\
        \bottomrule
\end{tabular}
\caption{Pruning performance of different datasets under unstructured 50\% sparsity on SparseGPT}
\label{tab:table3}
\end{table*}

\begin{table*}[t]
\centering
\small
\begin{tabular}{lllcccccc}
\toprule
        \textbf{Difficulty Level} & \textbf{GSM8K} & \textbf{MATH500} & \textbf{AIME} & \textbf{Olympiadbench} & \textbf{GPQA} & \textbf{AVG} & \textbf{Token Length}\\
        \midrule
        Easy & 91.3 & 88.9 & 30.8 & 49.6 & 30.8 & 58.3 & 2005\\
        Meidum & 91.0 & 90.5 & 35.4 & 50.3 & 32.3 & 59.9 & 3972\\
        Hard & 91.5 & 89.8 & 37.1 & 51.3 & 39.4 & 61.8 & 7375\\
        \bottomrule
\end{tabular}
\caption{Pruning performance using calibration data of varing difficulty levels under unstructured 50\% sparsity on SparseGPT}
\label{tab:table4}
\end{table*}

\appendix

\section{More Experiments on Qwen3-14B}

To further demonstrate the effectiveness of our method, we conducted experiments on the RL-based model Qwen3-14B.
As shown in Table~\ref{tab:qwen3}, SSGR consistently achieves the best performance, significantly outperforming C4 and MathR.

\begin{table*}[t]
\centering
\small
\begin{tabular}{lllcccccc}
\toprule
        \textbf{Qwen3-14B} & \textbf{GSM8K} & \textbf{MATH500} & \textbf{AIME} & \textbf{Olympiadbench} & \textbf{GPQA} & \textbf{AVG} \\
        \midrule
        C4 & 94.7 & 84.8 & 28.3 & 49.3 & 45.5 & 60.5 \\
        MathR & 94.6 & 92.4 & 51.8 & 57.6 & 47.0 & 68.7 \\
        SGR & 94.9 & 92.4 & 55.0 & 59.3 & 52.0 & 70.7 \\
        SSGR & 94.7 & 93.2 & 58.4 & 60.4 & 55.1 & 72.4 \\
        \bottomrule
\end{tabular}
\caption{Pruning performance of different datasets under unstructured 50\% sparsity on Qwen3-14B}
\label{tab:qwen3}
\end{table*}

\section{Ablation Study}

The results are shown in Table~\ref{tab:sl} and Table~\ref{tab:dt}.
As the sequence length increases, pruning performance generally improves.
However, performance at 16k is slightly lower than at 8k, which may be because the average length of the calibration dataset is closer to 8k, leading to better alignment and performance.
When no difficulty threshold is applied ($\tau$ = 1), pruning performance drops significantly.
Conversely, setting the threshold too low ($\tau$ = 0.25) leads to too few reasoning samples and insufficient diversity, which harms performance.
A moderate difficulty threshold ($\tau$ = 0.75 or 0.5) achieves the best results by balancing difficulty and diversity.

\begin{table*}[t]
\centering
\caption{Performance under different sequence lengths.}
\begin{tabular}{lcccccc}
\hline
\textbf{Sequence Length} & \textbf{0.5k} & \textbf{1k} & \textbf{2k} & \textbf{4k} & \textbf{8k} & \textbf{16k} \\
\hline
AVG & 60.0 & 60.6 & 60.8 & 60.8 & 62.5 & 62.0 \\
\hline
\end{tabular}
\label{tab:sl}
\end{table*}

\begin{table*}[t]
\centering
\caption{Performance under different difficulty thresholds.}
\begin{tabular}{lcccc}
\hline
\textbf{Difficulty Threshold} & 0.25 & 0.5 & 0.75 & 1 \\
\hline
\textbf{AVG} & 60.6 & 61.8 & 62.0 & 60.3 \\
\hline
\end{tabular}
\label{tab:dt}
\end{table*}

\section{More discussion on alternative CoT data generation methods}
As shown in Table~\ref{tab:md}, we generated reasoning data using Qwen3-14B (a stronger reasoning model). However, the pruning performance was inferior compared to data self-generated by DeepSeek-R1-Distill-Qwen-7B, particularly in terms of generalization, with a 10.1\% drop on GPQA. This further highlights the necessity of using self-generated reasoning data for effective pruning.

\begin{table*}[t]
\centering
\small
\begin{tabular}{lllcccccc}
\toprule
\textbf{Generation Source} & \textbf{GSM8K} & \textbf{MATH500} & \textbf{AIME} & \textbf{Olympiadbench} & \textbf{GPQA} & \textbf{AVG} \\
\midrule
DS-Distill-Qwen-7B & 91.1 & 89.9 & 39.2 & 51.1 & 38.8 & 62.0\\
Qwen3-14B & 89.7 & 87.9 & 35.8 & 49.4 & 28.7 & 58.3\\
\bottomrule
\end{tabular}
\caption{Pruning performance of different generation source.}
\label{tab:md}
\end{table*}

\end{document}